\documentclass{ifacconf}
\usepackage{mathtools}
\usepackage{amsmath}
\usepackage{amssymb}
\usepackage{graphicx}    
\usepackage{bm}

\usepackage{natbib}        
\usepackage{algorithm}
\usepackage{algpseudocode}
\usepackage{multicol}
\usepackage{subcaption}

\usepackage{xcolor}

\begin{document}
\begin{frontmatter}
\title{Towards Scalable Bayesian Optimization via Gradient-Informed Bayesian Neural Networks
} 

\thanks[footnoteinfo]{These authors contributed equally to this work. \\The work was supported by the U.S. National Science
Foundation under grants 2112754 and 2130734.}

\author[First]{Georgios Makrygiorgos\thanksref{footnoteinfo}}
\author[First]{Joshua Hang Sai Ip\thanksref{footnoteinfo}}
\author[First]{Ali Mesbah}

\address[First]{Department of Chemical and Biomolecular Engineering, University of California, Berkeley, CA 94720, USA.}

\begin{abstract} Bayesian optimization (BO) is a widely used method for data-driven optimization that generally relies on zeroth-order data of objective function to construct probabilistic surrogate models. These surrogates guide the exploration-exploitation process toward finding  global optimum. While Gaussian processes (GPs) are commonly employed as surrogates of the unknown objective function, recent studies have highlighted the potential of Bayesian neural networks (BNNs) as scalable and flexible alternatives. Moreover, incorporating gradient observations into GPs, when available, has been shown to improve BO performance. However, the use of gradients within BNN surrogates remains unexplored.
By leveraging automatic differentiation, gradient information can be seamlessly integrated into BNN training, resulting in more informative surrogates for BO. We propose a gradient-informed loss function for BNN training, effectively augmenting function observations with local gradient information. The effectiveness of this approach is demonstrated on well-known benchmarks in terms of improved BNN predictions and faster BO convergence as the number of decision variables increases. 
\end{abstract}


\end{frontmatter}

\section{Introduction}
Bayesian optimization (BO) \citep{movckus1975bayesian} has been a powerhouse for optimization of functions with unknown or partially known structure in diverse decision-making  applications. 
BO tackles black-box optimization
problems by sequentially solving easier optimization sub-problems. This is done by combining a probabilistic surrogate model, representing the objective as a function of decision inputs, and an acquisition function (AF) that, based on the current belief about the objective, suggests points in the design space to query the objective. 

BO is traditionally a zeroth-order optimization
method, exclusively relying on objective function observations to refine its belief of the objective function. As for surrogate modeling of the objective, Gaussian process (GP) regression is a popular choice \citep{williams2006gaussian}. Recent work has demonstrated the benefits of leveraging derivative information in BO via GPs; fully observed gradients can yield $D+1$ extra observations, where $D$ is the input dimensionality, per system query at a given input. This provides useful local information around the observed points, leading to a more accurate posterior for objective function approximation \citep{wu2017bayesian}.
Gradients have been shown to improve regret bounds in BO \citep{shekhar2021significance}, or used for combined global and local search in more complex, potentially multi-objective tasks
\citep{ip2025user, ip2024preference}. 
A standard approach of exploiting gradients (or higher-order derivatives) in BO with GPs is by defining a joint kernel that utilizes both function and gradient observations  \citep{eriksson2018scaling,wu2017exploiting}. Alternatively, to mitigate challenges due to noisy function and derivative observations,  gradients can be directly modeled using a probabilistic model and utilized to enforce optimality conditions in the AF \citep{makrygiorgos2022aegbo, makrygiorgos2023no}.

Nonetheless, GPs suffer from several limitations in this context, including slow inference times that scale cubically with the number of observations and input dimension $D$. In particular, the computational complexity of GP inference is $\mathcal{O}\left((D+1)^3n^3\right)$ and $\mathcal{O}\left((D+1)n^3\right)$ for joint and independent kernels, respectively \citep{wu2017bayesian}. This challenge has motivated probabilistic modeling using neural networks (NNs), while combining them with ideas from non-parametric kernel methods such as GPs.
Deep kernel learning \citep{wilson2016deep} is one such approach that improves the inference time of standard GPs from $\mathcal{O} (n^3)$ to $\mathcal{O} (n)$  by utilizing a NN to obtain a feature representation that is passed to a GP layer, enabling probabilistic inference.
Another class of models is neural processes that combines the flexibility of NNs with the probabilistic properties of stochastic processes \citep{garnelo2018neural}. Using an encoder-decoder framework with latent variables, neural processes provide a distribution over functions, capturing uncertainty in the latent representation.

In this work, we focus on Bayesian neural networks (BNNs), a class of probabilistic NNs that treat network parameters as random variables; hence, allowing for uncertainty quantification in predictions.  Dropout \citep{gal2016dropout}, which randomly inactivates weights during NN training, is one of the simplest mechanisms for uncertainty quantification of NNs. Another popular method relies on variational inference, which approximates the posterior over weights using the so-called Bayes by Backprop method by minimizing the divergence from the true posterior \citep{blundell2015weight, bao2023learning}. Unlike the aforementioned methods that typically lead to conservative and suboptimal uncertainty estimates, Hamiltonian Monte Carlo (HMC) is a sampling-based uncertainty quantification approach, where Hamiltonian dynamics are used to guide exploration of the posterior to propose new samples \citep{neal2012mcmc}. However, HMC does not scale favorably for handling a large number of uncertainties. In \citep{chen2014stochastic},  HMC is modified to address this limitation by using stochastic gradients, known as stochastic gradient HMC (SGHMC), and adding noise to maintain a consistent posterior distribution, making it practical for BNN training. Although SGHMC can be effective for probabilistic surrogate modeling in BO, custom design of the loss function is crucial for obtaining models that describe the general function shape in the $D$-dimensional input space. Similar to conditioning GPs on gradient observations that provide extra local information about the function, augmenting the loss function in BNN training with ``physically-relevant'' terms that can provide the same type of information is essential, especially in low-data problems where BO is usually used. 

We propose a new approach, termed gradient-informed neural network BO (GINNBO), that employs BNNs to learn scalable surrogates using both function and gradient observations. BNNs are $\mathcal{O}((D+1)n)$ in computational complexity, which makes them far more efficient than GPs, especially when gradients are involved in surrogate modeling. We introduce a weighted derivative-informed regularization term in the loss function, guiding the BNN training towards solutions that align the learned function with its gradient. This is achieved by incorporating a soft constraint negative log-likelihood (NLL) term based on gradient observations alongside the NLL term for function observations, enriching the loss function with local gradient information. As this only modifies the loss for BNN training, the proposed method is agnostic to the choice of AF.  
We demonstrate that incorporating gradients via independent surrogates is insufficient. Instead, we explicitly enforce the relationship between the function and its partial derivatives in the input search space via automatic differentiation. This is achieved via a joint gradient-informed loss that involves both the objective function and its partial derivatives.



\section{Preliminaries}

\subsection{Problem Statement}
We aim to minimize an expensive objective function $f(\mathbf{x})$ 
\begin{equation}
    \label{eq: minimize_objective}
    \mathbf x^* \in \arg\underset{\mathbf x}{\min}\ f(\mathbf x),
\end{equation}
where $\mathbf x \in \mathbb{R}^{D}$ are the decision variables and $D$ denotes the dimensionality of $\mathbf x$. We assume noisy observations of $f$ or $\nabla f$ are available, i.e.,
\begin{equation}
    \label{eq: noisy_observations}
    (y, \nabla y)^\top = (f(\mathbf x), \nabla f(\mathbf x))^\top + \bm \epsilon,
\end{equation}
where $\bm \epsilon \sim \mathcal{N}(\bm 0, \sigma_n^2I)$.

\subsection{Bayesian Neural Networks}

BO uses a surrogate model of the objective $f(\mathbf x)$. Here, we use BNNs due to favorable scalability, i.e., $\mathcal{O}((D+1)n)$.
BNNs extend traditional NNs by incorporating Bayesian inference, allowing the network to estimate the posterior distribution of its outputs with respect to its learnable parameters. 
We define a BNN surrogate of function $\phi(\mathbf x)$ as 
\begin{equation}
\label{proba_model}
p(\phi(\mathbf x) | \mathbf x,\theta) = \mathcal{N}(\hat{\phi}(\mathbf x,\theta_{\mu}), \theta_{\sigma^2}), 
\end{equation}
where
$\hat{\phi}(\mathbf x, \theta_{\mu}): \mathbb{R}^D \rightarrow \mathbb{R}$
is a parametric NN model evaluated for parameters $\theta_{\mu}$ and $\mathcal{N}$ denotes a normal distribution. As such, we denote the learnable parameters of a BNN model by $\theta=[\theta_{\mu},\theta_{\sigma^2}]^\top$.

Given a training dataset \(\mathcal{D} = \{(\mathbf{x}_i, y_i, \nabla y_i)\}_{i=1}^N\) with $N$ being the number of samples, a BNN learns the posterior distribution of parameters \(\theta\), i.e., \(p(\theta | \mathcal{D})\), rather than just finding a maximum likelihood estimate of $\theta$. Uncertainty in BNN predictions can then be derived by marginalizing over this posterior as
\begin{equation}
\label{marginalization}
p(\phi(\mathbf x) , \mathcal{D}) = \int p(\phi(\mathbf x) | \mathbf x,\theta) p(\theta | \mathcal{D}) \, d\theta.
\end{equation}
Furthermore, given the posterior parameter distribution  $ p(\theta |\mathcal{D} ) $, we can approximate the output posterior $p(\phi(\mathbf x), \mathcal{D})$ in \eqref{marginalization} in a sample-based manner to predict the mean $\mu(\cdot)$ and variance $\sigma^2(\cdot)$ of a BNN model as
\begin{subequations}
\label{eq: bnn_predictions}
\begin{align}
\label{eq: mean}
\mu\left( \mathbf{x} \right) &= \frac{1}{M} \sum_{i=1}^M \phi(\mathbf{x}; \theta^i_m), \\
\label{eq: var}
\sigma^2\left( \mathbf{x} \right) &= \frac{1}{M} \sum_{i=1}^M \left( \phi(\mathbf{x}; \theta^i) - \mu\left( \mathbf{x} \right) \right)^2 + \theta_{\sigma^2},
\end{align}
\end{subequations}
where $M$ denotes the number of samples $\theta^i \sim p(\theta |\mathcal{D} ) $ drawn from the posterior of $\theta$.

\subsection{Bayesian Optimization}

In BO, the next input query is designed by optimizing an AF, which realizes a tradeoff between \textit{exploration} and \textit{exploitation} of the input design space. Exploration seeks to explore regions that lack information about $f(\mathbf{x})$, towards searching for potentially better input designs. On the other hand, exploitation attempts to find better input designs in the vicinity of previously located desirable ones. Here, we use two AFs: lower confidence bound (LCB) and log expected improvement (LogEI). LCB is a commonly used AF in BO defined as \citep{srinivas2009gaussian}
\begin{equation}
\label{eq: LCB}
\text{LCB}(\mathbf x) = \mu(\mathbf x) - \beta \sigma(\mathbf x),
\end{equation}
where $\beta$ is a hyperparameter that balances the tradeoff between exploration and exploitation. LogEI is based on the popular expected improvement AF \citep{jones1998efficient}, where the logarithm yields approximately the same solutions while alleviating issues with numerical optimization due to vanishing gradients. LogEI is defined as \citep{ament2023unexpected}
\begin{equation}
    \label{eq: LogEI}
    \text{LogEI}(\mathbf x)=\log \left|h\left(\frac{\mu(\mathbf x)-y^*}{\sigma(\mathbf x)}\right)\right| + \log \left|\sigma(\mathbf x)\right|,
\end{equation}
where $y^*$ is the best $y$ found thus far,  $h(\cdot) = \phi_{G}(\cdot) + (\cdot)\Phi_{G}(\cdot)$, and $\phi_{G}$ and $\Phi_{G}$ denote the probability distribution function and cumulative distribution function of a normal distribution, respectively. 

Putting this together, a BO algorithm iteratively performs surrogate model training, AF optimization, and objective function query until the experimental or computational budget for querying the system is exhausted.

\section{Gradient-Informed Bayesian Neural Networks for Bayesian Optimization}

We now present the gradient-informed loss used for training a BNN surrogate, along with the proposed gradient-informed neural network BO (GINNBO) algorithm.

\subsection{Gradient-Informed Loss Function}

Here, we present a loss function that embeds gradient information in constructing a BNN surrogate.
Automatic differentiation allows for efficient computation of gradients. The gradients of a BNN output $\phi(\mathbf{x}; \theta^i)$ with respect to inputs $\mathbf{x}$ can be calculated by propagating the gradients backwards through the BNN using the chain rule 
\begin{equation}
\frac{\partial \phi(\mathbf{x}; \theta^i)}{\partial \mathbf{x}} = \frac{\partial \phi(\mathbf x;\theta^i)}{\partial z} \times \frac{\partial z}{\partial \mathbf{x}},
\end{equation}
where $z$ denotes the activation functions within the network. Let $\nabla_{\mathbf{x}} \phi(\mathbf{x}; \theta^i)$ denote the vector of all partial derivatives.
Similarly to the sample-based inference of the mean and variance of the BNN output in \eqref{eq: bnn_predictions}, we can approximate the mean and variance of gradients as 
\begin{subequations}
\label{eq: bnn_predictions_grad}
\begin{align}
\label{eq: mean_grad}
\mu_{\nabla}\left( \mathbf{x} \right) &= \frac{1}{M} \sum_{i=1}^M \nabla_{\mathbf{x}} \phi(\mathbf{x}; \theta^i), \\
\label{eq: var_grad}
\sigma^2_{\nabla}\left( \mathbf{x} \right) &= \frac{1}{M} \sum_{i=1}^M \left( \nabla_{\mathbf{x}} \phi(\mathbf{x}; \theta^i) - \mu_{\nabla}\left( \mathbf{x} \right) \right)^2 + \theta_{\sigma^2}.
\end{align}
\end{subequations}

We define the total loss for training a BNN surrogate in terms of two negative log-likelihood (NLL) loss functions. The first is the  NLL for fitting the BNN output, $\mathcal{L}_{f}$, defined as
\begin{equation} 
\label{eq: NLL}
\mathcal{L}_{f} = \frac{1}{N} \sum_{i=1}^N \left( \frac{(y_i - \mu(\mathbf{x}_i))^2}{2 \sigma^2(\mathbf{x}_i)} + \frac{1}{2} \log(2 \pi \sigma^2(\mathbf{x}_i)) \right).
\end{equation} 
$\mathcal{L}_{f}$ evaluates the likelihood of the predicted mean \(\mu(\mathbf{x}_i)\) and variance \(\sigma^2(\mathbf{x}_i)\) for each data point \(\mathbf{x}_i\), penalizing predictions that deviate from the observation \(y_i\) based on the assumption that the observation noise is Gaussian. As such, $\mathcal{L}_{f}$ accounts for both the predicted values and the associated uncertainty of  predictions. 

The second NLL loss is defined in terms of gradient predictions. The gradient loss  $\mathcal{L}_{\nabla f}$ takes the form of
\begin{equation}
\label{eq: NLL_grad}
\mathcal{L}_{\nabla f}  = \frac{1}{N} \sum_{i=1}^N \left( \frac{\|\nabla y_i - \mu_{\nabla}(\mathbf{x}_i)\|^2}{2 \sigma^2_{\nabla}(\mathbf{x}_i)} + \frac{1}{2} \log(2 \pi \sigma^2_{\nabla}(\mathbf{x}_i)) \right),
\end{equation}
where an $L_2$-norm is used since gradients comprise a vector. The gradient loss $\mathcal{L}_{\nabla f}$ can be viewed as a regularization term that applies the same NLL formulation as in \eqref{eq: NLL} to the partial gradients of the predicted mean. Specifically, \eqref{eq: NLL_grad} penalizes discrepancies between the true gradients \(\nabla y_i\) and the BNN predicted gradients with mean \( \mu_{\nabla}(\mathbf{x}_i)\) and variance \(\sigma^2_{\nabla}(\mathbf{x}_i)\).

By leveraging $\mathcal{L}_{\nabla f}$, we  ensure that the BNN surrogate would better align with the shape and slope of the target function, yielding a more accurate description of the unknown objective. Accordingly, we define the total loss function as
\begin{equation}
\label{eq: NLL_overall}
\mathcal{L} = \mathcal{L}_{f} + \lambda_{\nabla} \mathcal{L}_{\nabla f} + 
\mathcal{L_{\text{prior}}},
\end{equation}
where $\lambda_{\nabla}$ is a hyperparameter and $\mathcal{L}_{\text{prior}}$ is as a regularization term (e.g., Ridge regularization).
The total loss \eqref{eq: NLL_overall} makes the surrogate more sensitive to directional information in the input space.   $\lambda_{\nabla}$ scales the importance of the NLL of gradient observations in relation to function observations. $\lambda_{\nabla}$ should be selected such that the NLL losses $\mathcal{L}_{f}$ and $\mathcal{L}_{\nabla f}$ have the same order of magnitude, while it can be adjusted for noisy observations. 
In practice, $\lambda_{\nabla}$ can be set empirically via cross-validation.


\subsection{Stochastic Gradient Descent-Hamiltonian Monte Carlo}

Hamiltonian Monte Carlo (HMC) is a Markov Chain MC approach that uses the loss as a \emph{potential energy function} \( U(\theta) \), framing the problem of learning $\theta$ as a physical problem \citep{neal2012mcmc}.
To this end, HMC views the unknown parameters $\theta$ as ``position'' and introduces auxiliary variables \( \mathbf{r} \),  corresponding to  ``momentum" of a dynamical system. Accordingly, HMC looks to establish the joint distribution of $\theta$ and $\mathbf{r}$, i.e., 
\begin{equation}
\label{eq: joint_theta_r_D}
p(\theta, \mathbf{r} \mid \mathcal{D}) \propto \exp \left( -U(\theta) - \frac{1}{2} \mathbf{r}^T \mathbf{M}^{-1} \mathbf{r} \right),
\end{equation}
where $\mathbf{M}$ is a mass matrix modulating system dynamics. By simulating this Hamiltonian dynamical system, HMC generates samples according to \eqref{eq: joint_theta_r_D}, and the posterior distribution of $\theta$ is obtained by marginalizing over $\mathbf{r}$. 

In stochastic gradient HMC (SGHMC), the HMC approach is adapted for large-scale datasets by incorporating mini-batch gradients to alleviate the need for full-batch gradient computations and, thus, allow for efficient sampling  \citep{chen2014stochastic}. The updates of position $\theta$ and momentum $\mathbf{r}$ discretize Hamiltonian dynamics, while adding friction, via a matrix \( \mathbf{C} \), and noise terms to counterbalance the variance introduced by stochastic gradients. This ensures that the sampler remains stable and converges to the desired target distribution even in the presence of noisy updates.
To improve robustness in the sampling process, \citet{springenberg2016bayesian} propose several modifications to the standard SGHMC. These include adaptive choices for the mass matrix \( \mathbf{M} \), friction term \( \mathbf{C} \), and step size \( \delta \), making the method more robust to varying scales of parameters and noise levels in gradient estimates. The modified mass matrix \( \mathbf{M} \) is adapted based on an estimate of the gradient variance \( \hat{V}_{\theta} \), obtained via an exponential moving average of the gradient's element-wise variance during a ``burn-in" phase. The friction term \( \mathbf{C} \) is selected as \( \mathbf{C} = c \mathbf{I} \), where \( c \) is tuned relative to the step size \( \delta \) and the variance estimate \( \hat{V}_{\theta} \) to ensure that the system maintains stability under stochastic gradients. With these modifications, the update rules of the adaptive SGHMC approach used here become
\begin{subequations}
\begin{align}
\theta &\leftarrow \theta + \mathbf{v}, \\
\mathbf{v} &\leftarrow \mathbf{v} -\delta^2 \, \hat{V}_{\theta}^{-1/2} \nabla \tilde{U}(\theta) 
    - \delta \, \hat{V}_{\theta}^{-1/2} \mathbf{C} \, \mathbf{v} \notag \\
    &\quad + \mathcal{N} \left( 0, 2 \delta^3 \, \hat{V}_{\theta}^{-1/2} \mathbf{C} \, \hat{V}_{\theta}^{-1/2} 
    - \delta^4 \mathbf{I} \right),
\end{align}
\end{subequations}
where \( \mathbf{v} = \delta \hat{V}_{\theta}^{-1/2} \mathbf{r} \) is an auxiliary variable and \( \nabla \tilde{U}(\theta) \) denotes the stochastic gradient of the potential energy \( U(\theta) \) computed based on a mini-batch of data. Here, $\mathbf{v} $ stabilizes the updates by adapting to the variance structure of the parameter space, whereas the noise term \( \mathcal{N} \left( 0, 2 \delta^3 \, \hat{V}_{\theta}^{-1/2} \mathbf{C} \, \hat{V}_{\theta}^{-1/2} - \delta^4 \mathbf{I} \right) \) is dynamically scaled to counterbalance the variance introduced by mini-batch gradients to ensure the sampling robustness.
As such, these adaptive updates balance exploration and stability in training the BNN surrogate within each BO iteration by aligning the influence of each parameter's noise with its gradient variance. The proposed GINNBO approach is summarized in Algorithm~\ref{alg:GINNBO}.

\begin{algorithm}[tb]
\caption{Gradient-Informed Neural Network Bayesian Optimization (GINNBO)}
\label{alg:GINNBO}
\begin{algorithmic}[1]
\State Initialize dataset by obtaining $k$ initial observations $\mathcal{D} = \{(\mathbf{x}_i, y_i, \nabla y_i)\}_{i=1}^k$
\State Initialize hyperparameters for BNN $\phi(x;\theta^i)$; number of layers $N_{l}$, number of nodes per layer $N_{\text{nodes}}^l$, activation function $z$, and gradient regularization weight $\lambda_{\nabla}$
\State Initialize SGHMC parameters: total steps $N_{\text{MC}}$, ``Burn-in" steps $N_{\text{burn}}$, learning rate $\eta$, 
step size $\delta$, and sampling interval $w$
\While{computational budget not exceeded}
    \State \textbf{Train BNN:}
    \For{SGHMC step $s = 1$ to $N_{\text{MC}}$}
        \For{mini-batch $(\mathbf{x}_{\text{batch}}, y_{\text{batch}}, \nabla y_{\text{batch}})$ in $\mathcal{D}$}
            \State Calculate $\mathcal{L}_f$ and $\mathcal{L}_{\nabla f}$ for batch
            \State Compute total loss $\mathcal{L} = \mathcal{L}_f + \lambda_{\nabla} \mathcal{L}_{\nabla f} + \mathcal{L_{\text{prior}}}$
            \State Backpropagate $\mathcal{L}$ with SGHMC
        \EndFor
        \If{$s > N_{\text{burn}}$ \textbf{and} $(s - N_{\text{burn}}) \bmod w = 0$}
            \State Store current BNN weights sample
        \EndIf
    \EndFor
    \State \textbf{Optimize Acquisition Function (AF):}
    \State $\mathbf{x}' \gets \underset{\mathbf{x}}{\arg\min}\ \text{AF}(\mathbf{x})$
    \State New observation $(y', \nabla y')^\top = \left(f(\mathbf{x}'), \nabla f(\mathbf{x}')\right)^\top + {\mathbf{\epsilon}}$
    \State Update dataset $\mathcal{D} \leftarrow \mathcal{D} \cup \{(\mathbf{x}', y', \nabla y')\}$
\EndWhile
\end{algorithmic}
\end{algorithm}

\subsection{Illustrative Example of Effect of Gradient Information}

\begin{figure*}[t!]
    \centering
    \includegraphics[width=0.85\textwidth]{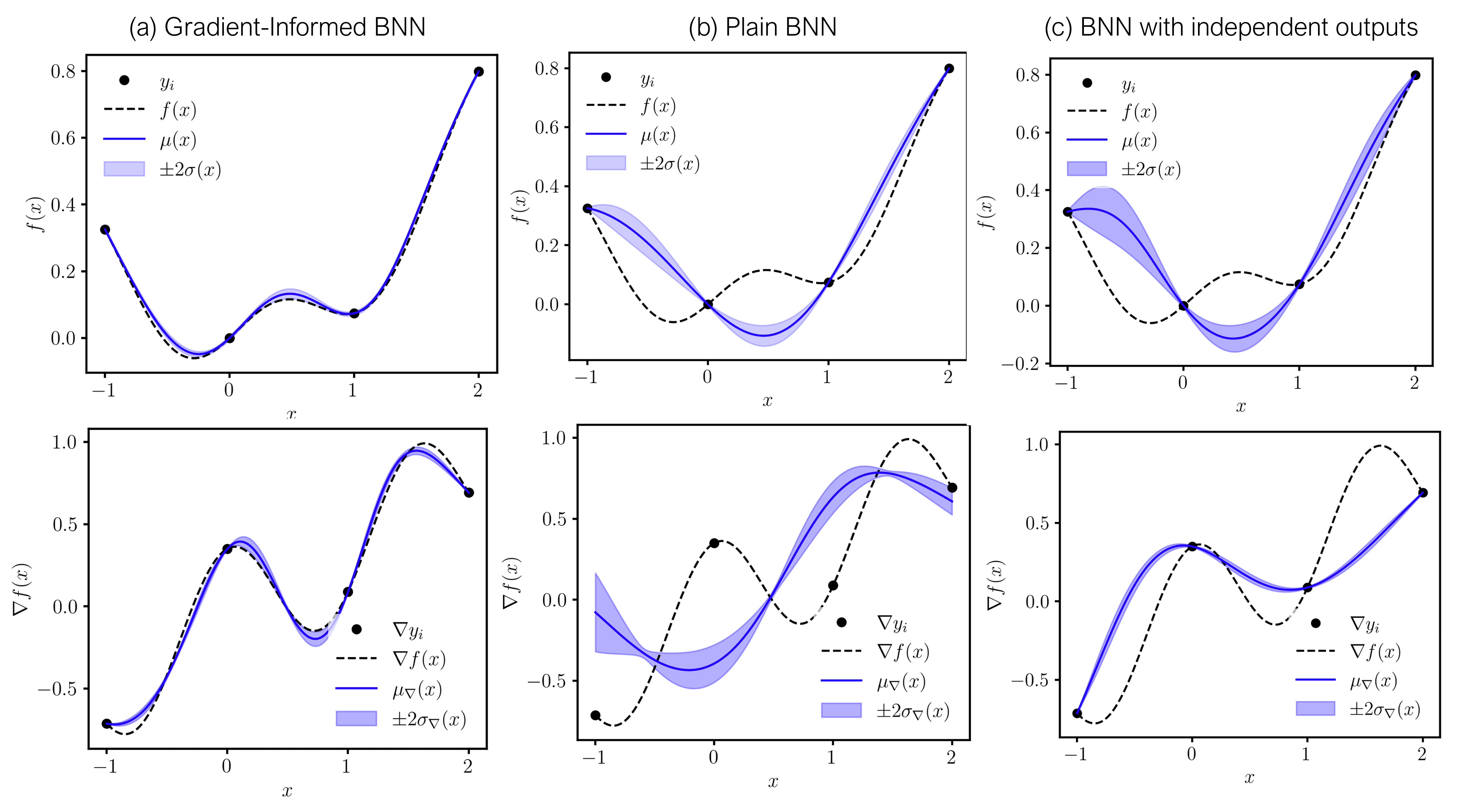}
    \caption{Effect of gradient information on predictive capability of a BNN surrogate. 
    (a) BNN trained with function and gradient observations using loss~\eqref{eq: NLL_overall}. (b) BNN trained with only function observations. (c) BNN trained with function and gradient observations, but treated as independent outputs of the BNN.
    The blue line represents the predicted mean, while the blue shaded region indicates the uncertainty (±2 standard deviations) around the prediction. In cases (a) and (b), derivative is computed via automatic differentiation of the BNN output.} 
    \label{fig:1d_demo}
\end{figure*}

To demonstrate the effectiveness of the gradient-informed total loss~\eqref{eq: NLL_overall} on the predictive capability of a BNN surrogate, we consider a 1D function $f(x)$, taken from \citep{paulson2022adversarially}. This function exhibits two local minima. In Fig. \ref{fig:1d_demo}, we present predictions of three different BNN models: BNN trained with the proposed loss \eqref{eq: NLL_overall} that captures the function and its derivative jointly (Fig. \ref{fig:1d_demo}a); BNN trained with only function observations (Fig. \ref{fig:1d_demo}b); and BNN trained to predict both the function and its derivative, i.e., 
$\hat{\boldsymbol{\phi}}(x, \theta_{\mu}): \mathbb{R}^D \rightarrow \mathbb{R}^{D+1}$ (Fig. \ref{fig:1d_demo}c).
In the latter case, the loss is \eqref{eq: NLL_overall}; however, the posterior of gradients in \eqref{eq: mean_grad} and \eqref{eq: var_grad}
are evaluated using the $i^{th}$ BNN output $\phi_i(x,\theta_{\mu}), \, i=1,\dots,D+1$. 
In Fig. \ref{fig:1d_demo}b, when only function observations are used as training data, the BNN predicts a function with a parabolic shape, completely missing the existence of the two local minima. In this case, the gradient of the BNN output, computed via automatic differentiation, deviates significantly from the true derivative. In Fig. \ref{fig:1d_demo}c, 
SGHMC converges to inference mean and variance 
that minimize the NLL loss independently for each output, without enforcing an explicit connection between them. Therefore, although both the function and derivative predictions interpolate between the training samples, they fail to capture the true shape of the function. More importantly, the function and gradient predictions do not match the real values.

In contrast, when derivative observations are  incorporated into the total loss \eqref{eq: NLL_overall}, while the gradients are computed via automatic differentiation and not as independent outputs as in Fig. \ref{fig:1d_demo}c, the BNN surrogate captures the true function almost perfectly (Fig. \ref{fig:1d_demo}a). This is because the BNN learns not only the function at the training samples, but also the directionality of the function. Notably, in all cases, SGHMC offers a reliable characterization of uncertainty, exhibiting very low uncertainty around observed samples and reasonable bounds in unexplored regions. This reliable uncertainty estimation may not be achieved with other methods, such as variational inference.

\section{Case Studies}

We demonstrate the performance of the proposed GINNBO approach on three benchmark problems of increasing dimensions and complexity, namely the McCormick (2D) \citep{adorio2005mvf}, Rosenbrock (4D) \citep{picheny2013benchmark}, and Hartmann (6D) \citep{dixon1978global} functions. The performance of GINNBO with the total loss \eqref{eq: NLL_overall} (with $\lambda_{\nabla}=1$) is compared with the baseline case without the gradient NLL loss, i.e., $\lambda_{\nabla}=0$ in \eqref{eq: NLL_overall}. The AFs used are LCB \eqref{eq: LCB} and LogEI \eqref{eq: LogEI}. Since the analytical expressions for the three benchmark problems are known, their exact derivatives can be computed. Here, we consider the case of noiseless gradient observations. In all cases, the initial BNN surrogate is learned using $2D$ samples of $\mathbf{x}$ selected randomly. For all BNN surrogates, we use $N_\text{l}=5$ hidden layers with $N^\text{l}_\text{nodes}=80$ nodes and \texttt{tanh} activation function. For the SGHMC algorithm, we use approximately $N_\text{MC}=6000$ steps with a burn-in phase of $N_\text{burn}=2000$. In the implementation of the AF optimization, L-BFGS with restarts is used.

\begin{figure*}[t!]
    \centering
    \includegraphics[width=1\textwidth]{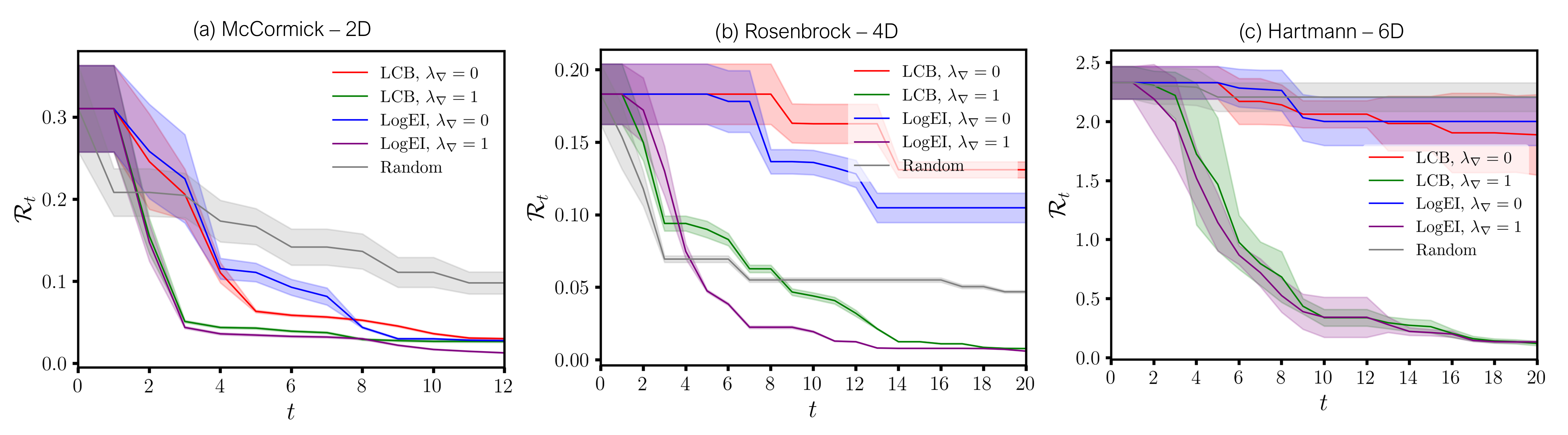}
    \caption{Instantaneous regret plots for (a) Mccormick (2D), (b) Rosenbrock (4D), and (c) Hartmann (6D) benchmark problems. Comparisons are demonstrated for the base case (with no gradient NLL term in the loss function \eqref{eq: NLL_overall}, $\lambda_{\nabla}=0$) and the case of the gradient NLL term with $\lambda_{\nabla}=1$ for the LCB and LogEI acquisition functions. Each experiment is performed with 10 random seeds and the mean $\pm$ 1 standard deviation are displayed.}
    \label{fig:BO_results}
\end{figure*}

We define $\mathcal{R}_t$ as the instantaneous regret at iteration $t$, i.e., $\mathcal{R}_t=f(\mathbf{x}^*) - \mathcal{F}(x_t) $, where $\mathcal{F}(x_t)$ is the cumulative best observation (incumbent best) at iteration $t$, as our performance metric. Fig.~\ref{fig:BO_results} shows the convergence plots of GINNBO and the baseline algorithm with $\lambda_{\nabla}=0$ for the three benchmark problems. All regret profiles are based on the mean $\pm 1$ standard deviation of $\mathcal{R}_t$ over 10 random seeds.  Fig.~\ref{fig:BO_results} suggests that GINNBO of Algorithm~\ref{alg:GINNBO} outperforms the baseline BO algorithm for both LCB and LogEI AFs across all three benchmarks. It is evident that GINNBO is agnostic to the choice of AF. 
In all cases, GINNBO with $\lambda_{\nabla}=1$ in the total loss function reaches much smaller values of the instantaneous regret $\mathcal{R}_t$ at earlier iteration indices $t$. When gradient information is available, the BNN surrogate is more likely to locate regions that may contain the global minimum, but can also provide more reliable predictions of the unknown objective (as also seen in Fig. \ref{fig:1d_demo}), in turn enabling less redundant exploration.  
The extent to which GINNBO outperforms the baseline algorithm varies with the problem dimensionality $D$. The improved performance is more pronounced in higher dimensions. 
Even in lower dimensions, it becomes more challenging to learn an accurate BNN surrogate in the limit of sample size.
Gradient information allows for effectively scaling the available information per each training sample, $D$-fold; hence, enabling more accurate models to be trained. 
As can be seen, data scarcity in the case of no derivative observations can make the BNN-based search comparable (or worse) to that of random search. 


 

\begin{table}[b!]
\centering
\renewcommand{\arraystretch}{1.2} 
\begin{tabular}{|c|c|c|c|}
\hline
\textbf{Benchmark problem} & $\boldsymbol D$ & \textbf{Mean $\pm$ Std. Dev. (s)} \\
\hline
Mccormick & 2 & 23.09 $\pm$ 1.26 \\
\hline
Rosenbrock & 4 & 23.36 $\pm$ 0.52\\
\hline
Hartmann & 6 & 23.71 $\pm$ 0.29\\
\hline
\end{tabular}
\caption{Computational times for training a BNN surrogate with the gradient-informed loss \eqref{eq: NLL_overall} 
for the McCormick (2D), Rosenbrock (4D), and Hartmann (6D) problems.}
\label{tab:results}
\end{table}

Furthermore, we investigate the computational complexity of GINNBO with the LCB AF in relation to dimension $D$. Note that GINNBO is agnostic to AF used. As shown in Table~\ref{tab:results}, the computational times for training the BNN surrogate are nearly identical across all benchmark problem under identical SGHMC settings. This is expected since BNNs are scalable due to their NN architecture, which makes them an attractive option for higher-dimensional problems where traditional BO with GPs becomes computationally inefficient. Here, at each BO iteration, a BNN surrogate is re-trained from scratch based on the available dataset  $\mathcal{D}_t$. It should be noted that transfer learning techniques can further accelerate the time needed to train each model. 


\section{Conclusion}

This paper presented a gradient-informed Bayesian neural network BO (GINNBO) approach, as a scalable alternative for BO when gradient information is available.
GINNBO uses a gradient-based negative log-likelihood term in the loss function, acting as a regularization term, to learn a more accurate model of the blackbox objective function in BO. Numerical experiments on several benchmark problems demonstrate that GINNBO consistently outperforms the case in which the BNN surrogate is only trained using function observations. We also demonstrate that the computational complexity of the BNN is independent of the dimensionality of the input space, making it an attractive alternative for gradient-informed BO compared to GP-based methods. Future work will investigate the effectiveness of GINNBO on real-life applications in comparison with alternative gradient-informed GP-based BO methods. We will also investigate the effect of BNN and BO hyperparameters on GINNBO performance.



\bibliography{ifacconf}             
                                                   







\end{document}